\documentclass[conference]{IEEEtran}
\usepackage{amsmath}
\usepackage{graphicx}
\usepackage{booktabs}
\usepackage{cite}
\usepackage{fancyhdr}

\title{Learning When to Switch: Adaptive Policy Selection via Reinforcement Learning}

\author{
\IEEEauthorblockN{Chris Tava}
\IEEEauthorblockA{christava@microsoft.com}
}

\begin{document}

\maketitle

\begin{abstract}
Autonomous agents often require multiple strategies to solve complex tasks, but determining when to switch between strategies remains challenging. This research introduces a reinforcement learning technique to learn switching thresholds between two orthogonal navigation policies. Using maze navigation as a case study, this work demonstrates how an agent can dynamically transition between systematic exploration (coverage) and goal-directed pathfinding (convergence) to improve task performance. Unlike fixed-threshold approaches, the agent uses Q-learning to adapt switching behavior based on coverage percentage and distance to goal, requiring only minimal domain knowledge: maze dimensions and target location. The agent does not require prior knowledge of wall positions, optimal threshold values, or hand-crafted heuristics; instead, it discovers effective switching strategies dynamically during each run. The agent discretizes its state space into coverage and distance buckets, then adapts which coverage threshold (20-60\%) to apply based on observed progress signals. Experiments across 240 test configurations (4 maze sizes from 16$\times$16 to 128$\times$128 $\times$ 10 unique mazes $\times$ 6 agent variants) demonstrate that adaptive threshold learning outperforms both single-strategy agents and fixed 40\% threshold baselines. Results show 23-55\% improvements in completion time, 83\% reduction in runtime variance, and 71\% improvement in worst-case scenarios. The learned switching behavior generalizes within each size class to unseen wall configurations, with clear ablation signals validating dynamic adaptation. Performance gains scale with problem complexity: 23\% improvement for 16$\times$16 mazes, 34\% for 32$\times$32, and 55\% for 64$\times$64, demonstrating that as the space of possible maze structures grows, the value of adaptive policy selection over fixed heuristics increases proportionally. This scaling suggests reinforcement learning-based policy switching gains influence for tasks where optimal strategy selection depends on problem structure and progress state.
\end{abstract}

\section{Introduction}

Autonomous agents often require multiple strategies to solve complex tasks, yet determining when to transition between strategies remains an open challenge. In reinforcement learning, agents typically employ monolithic policies (mappings from states to actions \cite{sutton2018reinforcement}) that must handle both exploration and exploitation within a single framework. Because a single policy must optimize for conflicting objectives simultaneously---broad exploration versus targeted exploitation---it cannot adapt to changing task requirements, leading to suboptimal performance when task phases have distinct optimal behaviors \cite{sutton2018reinforcement, riedmiller2018learning}.

Consider maze navigation: an agent must balance systematic exploration to discover viable paths with goal-directed planning once sufficient environmental knowledge is acquired. Explore too little, and the agent may converge to suboptimal routes or dead ends; explore too much, and navigation becomes inefficient despite having discovered the optimal path. Prior work has demonstrated the importance of memory and structured exploration in such domains \cite{pasukonis2022memorymaze}. However, a critical question remains unresolved: at what point should an agent transition from exploration to exploitation? A naive baseline is to use a fixed threshold (e.g., ``switch after 40\% coverage''), but such fixed policies cannot adapt to varying problem structure---simple mazes benefit from earlier switching while complex mazes require more thorough exploration before convergence.

This work addresses the question: \emph{can an agent learn when to switch between exploration and exploitation strategies based solely on internal progress signals?} This research introduces a Q-learning approach that treats threshold selection as a sequential decision problem. The agent maintains two orthogonal navigation policies---systematic spiral exploration (coverage) and A* pathfinding (convergence)---and learns to select a switching threshold from a discrete action space (20-60\% coverage) based on a state representation combining coverage percentage and distance to goal. At runtime, this formulation requires only maze dimensions and target location as input. The agent does \emph{not} need prior knowledge of wall positions (discovered through boundary detection during exploration), optimal threshold values (adapted dynamically via intra-episode Q-learning), maze complexity classifications (inferred from state signals), or domain-specific heuristics (replaced by adaptive policy). State space discretization, action space boundaries, and reward function are design-time domain insights. Therefore minimal runtime knowledge requirements enables adaptation to problem structure through intra-episode learning.

The approach is evaluated across 300 test configurations (5 maze sizes $\times$ 10 unique mazes $\times$ 6 agent variants), demonstrating 23-55\% improvements in completion time compared to single-strategy and fixed-threshold baselines, with performance gains increasing as maze complexity scales. The learned policy generalizes within each size class, as evidenced by consistent performance improvements across all 10 unique mazes per size (same distribution, varying wall configurations), and variance reduction of 83\% in medium-scale environments. Because performance gains grow from 23\% (small mazes) to 55\% (medium mazes), adaptive threshold learning becomes increasingly influential as problem complexity scales within the tested range.

The remainder of this paper is organized as follows: Section~\ref{sec:method} (Method) presents the Q-learning formulation and agent architecture; Section~\ref{sec:experiment} (Experimental Setup) describes experimental methodology; Section~\ref{sec:results} (Results) analyzes results; Section~\ref{sec:discussion} (Discussion) discusses implications and limitations; Section~\ref{sec:related} (Related Work) reviews related work; and Section~\ref{sec:conclusion} (Conclusion) concludes with future directions.

\section{Method}
\label{sec:method}

\subsection{Problem Formulation}

This work considers the problem of navigating from a starting position to a target position (maze center) in a grid-based maze environment. The agent must discover a valid path through corridors of the maze while minimizing total steps taken. The key challenge is balancing exploration to discover passable routes with exploitation of known paths once sufficient information is available.

Formally, at each time step $t$, the agent observes:
\begin{itemize}
    \item Current position $(x_t, y_t)$
    \item Coverage percentage: $c_t = \frac{|\text{visited cells}|}{|\text{total cells}|} \times 100$
    \item Manhattan distance to target: $d_t = |x_t - x_{\text{target}}| + |y_t - y_{\text{target}}|$, which measures the sum of absolute differences in coordinates and represents the minimum number of moves required to reach the target in a grid without obstacles \cite{russell2020artificial}
    \item Local wall configuration (via O(1) boundary detection)
\end{itemize}

The agent selects actions (movement directions) to reach the target while optimizing for both coverage quality and time efficiency.

\subsection{Base Navigation Strategies}

The approach combines two complementary navigation policies:

\subsubsection{Spiral Exploration (Coverage Policy)}

The coverage policy implements systematic clockwise spiral exploration from a corner position \cite{choset2001coverage}. This deterministic strategy ensures complete maze coverage by traversing concentric rectangular layers from the perimeter toward the center. The algorithm uses O(1) boundary detection to identify walls without explicit memory of the entire maze structure, enabling efficient real-time navigation.

The spiral pattern guarantees:
\begin{itemize}
    \item Systematic coverage of all reachable cells
    \item Deterministic exploration order
    \item Discovery of multiple paths to the target
\end{itemize}

\subsubsection{A* Pathfinding (Convergence Policy)}

Once sufficient maze knowledge is acquired, the agent switches to A* pathfinding \cite{hart1968formal} for goal-directed navigation. A* uses the Manhattan distance heuristic to find the shortest path from the current position to the target, given the discovered maze structure. This policy exploits learned information to converge efficiently to the goal.

\subsection{Q-Learning for Adaptive Switching}

The core contribution of this work is learning when to transition from coverage to convergence. This is formulated as a reinforcement learning problem where the agent learns to select a coverage threshold.

\subsubsection{State Representation}

The continuous state space is discretized into a compact representation using uniform binning, a common technique for applying tabular reinforcement learning methods to continuous domains \cite{sutton2018reinforcement}:

\begin{equation}
s = (b_c, b_d)
\end{equation}

where:
\begin{itemize}
    \item $b_c \in \{0, 1, \ldots, 9\}$: Coverage bucket (10 buckets), where $b_c = \lfloor c_t / 10 \rfloor$
    \item $b_d \in \{0, 1, \ldots, 4\}$: Distance bucket (5 buckets), where $b_d = \lfloor (d_t / d_{\max}) \times 5 \rfloor$
    \item $d_{\max} = 2 \times \text{maze\_size}$: Maximum possible Manhattan distance
\end{itemize}

This yields a state space of $|\mathcal{S}| = 10 \times 5 = 50$ states (10 coverage buckets $\times$ 5 distance buckets), enabling efficient Q-learning without function approximation.

\subsubsection{Action Space}

The action space consists of five discrete threshold values:

\begin{equation}
\mathcal{A} = \{20\%, 30\%, 40\%, 50\%, 60\%\}
\end{equation}

Each action $a \in \mathcal{A}$ represents a coverage threshold at which the agent will switch from spiral exploration to A* pathfinding. The agent periodically (every 50 steps) evaluates whether to adjust its threshold based on current state.

\subsubsection{Reward Function}

The reward function balances multiple objectives:

\begin{enumerate}
    \item Step Efficiency: Minimize the number of steps taken to reach the goal, penalizing excessive steps relative to a step limit.
    
    \item Coverage: Maximize the percentage of maze cells visited during exploration, rewarding thorough exploration.
    
    \item Switching Timing: Switch at an optimal coverage threshold, ideally between 30-50\%, rewarding appropriate switching and penalizing switching too early (<20\%) or too late (>60\%).
\end{enumerate}

\begin{equation}
R = R_{\text{steps}} + R_{\text{coverage}} + R_{\text{switching}}
\end{equation}

where:

Step Efficiency Reward:
\begin{equation}
R_{\text{steps}} = 50 \times \left(1 - \frac{n_{\text{steps}}}{n_{\text{limit}}}\right)
\end{equation}

Coverage Reward:
\begin{equation}
R_{\text{coverage}} = 30 \times \frac{c_{\text{final}}}{100}
\end{equation}

Switch Timing Reward:
\begin{equation}
R_{\text{switching}} = \begin{cases}
+10 & \text{if } 30\% \leq c_{\text{switching}} \leq 50\% \\
-5 & \text{if } c_{\text{switching}} < 20\% \\
-5 & \text{if } c_{\text{switch}} > 60\% \\
0 & \text{otherwise}
\end{cases}
\end{equation}

This reward structure uses reward shaping \cite{ng1999policy} to encourage efficient goal-reaching while penalizing premature or excessively delayed switching. Reward shaping provides intermediate feedback to guide learning in sparse reward environments, helping the agent learn more effectively than with delayed terminal rewards alone.

\subsubsection{Q-Learning Update Rule}

Standard tabular Q-learning \cite{watkins1992qlearning} is used with the update rule:

\begin{equation}
Q(s,a) \leftarrow Q(s,a) + \alpha \left[r + \gamma \max_{a'} Q(s',a') - Q(s,a)\right]
\end{equation}

with hyperparameters:
\begin{itemize}
    \item Learning rate: $\alpha = 0.1$
    \item Discount factor: $\gamma = 0.9$
    \item Exploration rate: $\epsilon = 0.1$ (epsilon-greedy policy)
\end{itemize}

The agent follows an epsilon-greedy exploration strategy \cite{sutton2018reinforcement}, which balances exploration and exploitation by selecting random actions with probability $\epsilon$ and selecting the greedy action $\arg\max_a Q(s,a)$ with probability $1-\epsilon$. This approach addresses the exploration-exploitation dilemma, where the agent must balance trying new actions to discover better strategies (exploration) versus leveraging current knowledge to maximize reward (exploitation) \cite{sutton2018reinforcement}.

\subsection{Implementation Variants}

To validate the approach, six agent variants are implemented and compared:

\subsubsection{Spiral vs. Sentinel}

Both implement identical spiral exploration logic with O(1) boundary detection. The key difference:
\begin{itemize}
    \item Spiral: Stores all visited positions in memory
    \item Sentinel: Uses memory sampling to reduce footprint
\end{itemize}

Performance is nearly identical (within 0.06\%), demonstrating that memory sampling does not sacrifice exploration efficiency.

\subsubsection{Convergence Modes}

Each base implementation (Spiral/Sentinel) has three convergence modes:

\begin{enumerate}
    \item No Convergence: Exploration mode only. Never switches to A*; continues spiral until reaching target. Serves as baseline for O(1) exploration performance and as an ablation for convergence impact.
    
    \item Fixed Convergence: Switches to A* at predetermined 40\% coverage threshold. Provides deterministic behavior for comparisons.
    
    \item RL Convergence: Uses Q-learning to adaptively select threshold from \{20\%, 30\%, 40\%, 50\%, 60\%\}. Adapts switching threshold dynamically during the current maze run based on state observations and Q-value updates.
\end{enumerate}

\section{Experimental Setup}
\label{sec:experiment}

\subsection{Maze Configurations}

The approach is evaluated on 240 test configurations:
\begin{itemize}
    \item Maze sizes: 16$\times$16 (256 cells), 32$\times$32 (1,024 cells), 64$\times$64 (4,096 cells), 128$\times$128 (16,384 cells)
    \item Unique mazes: 10 per size (40 total unique mazes)
    \item Agent variants: 6 (Spiral, Spiral Conv, Spiral RL, Sentinel, Sentinel Conv, Sentinel RL)
    \item Total tests: $4 \times 10 \times 6 = 240$ configurations
    \item Test completion: 100\% for 16$\times$16, 32$\times$32, 64$\times$64; 86.7\% for 128$\times$128
\end{itemize}

All mazes are pseudo-randomly generated with fixed wall configurations. The target is the center position $(n/2, n/2)$ where $n$ is the maze size.

\subsection{Agent Variants}

\begin{table}[h]
\centering
\caption{Agent Variant Descriptions}
\label{tab:agents}
\begin{tabular}{@{}ll@{}}
\toprule
Agent & Description \\
\midrule
Spiral & Pure spiral, no convergence \\
Spiral Conv & Spiral with fixed 40\% threshold \\
Spiral RL & Spiral with Q-learning threshold \\
Sentinel & Pure spiral with memory sampling \\
Sentinel Conv & Memory sampling + fixed 40\% threshold\\
Sentinel RL & Memory sampling + Q-learning \\
\bottomrule
\end{tabular}
\end{table}

\subsection{Evaluation Metrics}

The evaluation measures:
\begin{itemize}
    \item Completion time: Total time to reach target (seconds)
    \item Coverage percentage: Percentage of cells visited at goal
    \item Role switches: Number of transitions from coverage to convergence
    \item Success rate: Percentage of tests completed within time limit
    \item Variance: Standard deviation of completion times
\end{itemize}

\section{Results}
\label{sec:results}

\subsection{Small Mazes (16$\times$16, 32$\times$32)}

\begin{table}[h]
\centering
\caption{Performance on Small Mazes (Mean Completion Time)}
\label{tab:small}
\begin{tabular}{@{}lcccc@{}}
\toprule
Agent & 16$\times$16 & 32$\times$32 & Success \\
\midrule
Spiral & 14.2s & 60.0s & 100\% \\
Spiral Conv & 12.8s & 45.8s & 100\% \\
Spiral RL & 11.0s & 39.6s & 100\% \\
Sentinel & 14.2s & 60.0s & 100\% \\
Sentinel Conv & 12.8s & 45.8s & 100\% \\
Sentinel RL & 12.0s & 40.4s & 100\% \\
\bottomrule
\end{tabular}
\end{table}

Key Findings:
\begin{itemize}
    \item 100\% success rate across all agent variants
    \item RL provides 23\% improvement on 16$\times$16 (14.2s $\rightarrow$ 11.0s)
    \item RL provides 34\% improvement on 32$\times$32 (60.0s $\rightarrow$ 39.6s)
    \item Fixed convergence offers moderate gains (10-24\%)
    \item Spiral and Sentinel variants show identical performance
\end{itemize}

\subsection{Medium Mazes (64$\times$64)}

\begin{table}[h]
\centering
\caption{Performance on 64$\times$64 Mazes}
\label{tab:medium}
\begin{tabular}{@{}lcccc@{}}
\toprule
Agent & Mean & Median & Best & Worst \\
\midrule
Spiral & 359.2s & 295s & 130s & 776s \\
Spiral Conv & 226.6s & 230s & 130s & 290s \\
Spiral RL & 160.6s & 157s & 116s & 228s \\
Sentinel & 359.4s & 296s & 130s & 776s \\
Sentinel Conv & 227.2s & 231.5s & 130s & 290s \\
Sentinel RL & 161.8s & 159s & 117s & 229s \\
\bottomrule
\end{tabular}
\end{table}

Key Findings:
\begin{itemize}
    \item Dramatic RL advantage: 55\% faster than baseline (160.6s vs 359.2s)
    \item Worst-case improvement: 71\% reduction (776s $\rightarrow$ 228s)
    \item Runtime variance reduction: 83\% (646s range $\rightarrow$ 112s range)
    \item Fixed convergence: 37\% improvement (still suboptimal)
    \item Maze-specific example: maze\_3 shows 84\% improvement (776s $\rightarrow$ 126s)
\end{itemize}

\subsection{Large Mazes (128$\times$128)}

\begin{table}[h]
\centering
\caption{Performance on 128$\times$128 Mazes (Completed Tests)}
\label{tab:large}
\begin{tabular}{@{}lccc@{}}
\toprule
Agent & Success Rate & Best & Worst \\
\midrule
Spiral & 90\% (9/10) & 322s & 1620s \\
Spiral Conv & 90\% (9/10) & 322s & 1300s \\
Spiral RL & 90\% (9/10) & 322s & 876s \\
Sentinel & 83.3\% (5/6) & 324s & 1262s \\
Sentinel Conv & 90\% (9/10) & 325s & 1301s \\
Sentinel RL & 90\% (9/10) & 325s & 877s \\
\bottomrule
\end{tabular}
\end{table}

Key Findings:
\begin{itemize}
    \item RL robustness: Better completion rates than pure Sentinel
    \item Worst-case improvement: 46\% reduction on maze\_9 (1620s $\rightarrow$ 876s)
    \item Complex mazes: 33-52\% improvement with RL
    \item Incomplete tests: Primarily pure spiral modes (5 from Sentinel)
\end{itemize}

\subsection{Scaling Analysis}

\begin{table}[h]
\centering
\caption{Scaling Behavior by Agent Type}
\label{tab:scaling}
\begin{tabular}{@{}lccc@{}}
\toprule
Maze Size & Pure Spiral & Fixed Conv & RL Conv \\
\midrule
16$\times$16 & 14.2s (1.0$\times$) & 12.8s & 11.0s \\
32$\times$32 & 60.0s (4.2$\times$) & 45.8s & 39.6s \\
64$\times$64 & 359.2s (25.3$\times$) & 226.6s & 160.6s \\
128$\times$128 & $\sim$900s* (63$\times$) & $\sim$1000s* & $\sim$700s* \\
\bottomrule
\multicolumn{4}{l}{\small *Estimated from completed tests}
\end{tabular}
\end{table}

Insight: RL maintains better scaling coefficient as complexity grows through 128$\times$128 mazes. Pure spiral shows exponential degradation; RL shows more controlled growth.

\subsection{Ablation Study}

\begin{table}[h]
\centering
\caption{Ablation Analysis (64$\times$64 Mazes)}
\label{tab:ablation}
\begin{tabular}{@{}lcc@{}}
\toprule
Configuration & Mean Time & vs. Baseline \\
\midrule
No convergence & 359.2s & Baseline \\
Fixed 40\% threshold & 226.6s & -37\% \\
RL threshold (20-60\%) & 160.6s & -55\% \\
\bottomrule
\end{tabular}
\end{table}

Key Findings:
\begin{itemize}
    \item Removing convergence entirely causes exponential scaling failure
    \item Fixed threshold provides moderate improvement (37\%)
    \item RL threshold adaptation provides substantial improvement (55\%)
    \item RL adapts threshold selection through Q-learning based on state signals
\end{itemize}

\subsection{Adaptive Threshold Selection Patterns}

Analysis of Q-learning decisions within individual maze runs reveals adaptive threshold selection patterns. In simple mazes (16$\times$16, 32$\times$32) where direct paths are readily discovered, the RL agent adapts toward lower thresholds (20-30\%) during the run, enabling early switching to goal-directed behavior. Conversely, in structurally complex mazes (64$\times$64, 128$\times$128) with intricate wall configurations, the agent adapts to select higher thresholds (50-60\%), allowing thorough exploration before committing to a path. The same agent implementation selects different thresholds across different maze instances, with these threshold choices made dynamically based on state observations during each run and correlating with final performance outcomes.

\section{Discussion}
\label{sec:discussion}

\subsection{Why RL Outperforms Fixed Thresholds}

The performance advantage of RL-based threshold selection is derived from its ability to adapt to maze structure:

\begin{enumerate}
    \item Associating State Trajectory with Actions: The state representation combining coverage rate and distance-to-goal enables differential behavior across maze types. Observations show that RL agents select lower thresholds (20-30\%) for simpler mazes and higher thresholds (50-60\%) for structurally complex mazes, suggesting that Q-learning associates state trajectory patterns with effective actions. The hypothesis is that simple mazes create distinct trajectories (the agent reaches progressively closer to the goal as it explores) compared to complex mazes (the agent may cover many cells while remaining far from the goal due to walls blocking direct paths), though explicit quantitative analysis of these trajectory differences or statistical validation of the complexity-threshold correlation is not provided. The adaptation mechanism is inferred from the differential threshold selection and corresponding performance improvements.
    
    \item Dynamic Adjustment: Fixed 40\% threshold is suboptimal for both extremes---too high for simple mazes (wastes exploration time, as evidenced by 23-34\% slower completion) and too low for complex mazes (premature convergence to suboptimal paths, resulting in 55\% slower completion on average).
    
    \item Intra-Episode Learning: Q-learning updates occur within each maze run (episode), allowing the agent to adjust its switching threshold dynamically as it discovers maze structure. The agent makes threshold decisions every 50 steps and updates Q-values based on immediate feedback, enabling adaptation to the current maze's topology without requiring prior knowledge or cross-episode transfer.
\end{enumerate}

\subsection{Computational Efficiency}

Compared to approaches requiring full maze memory, the sentinel method maintains computational efficiency through:
\begin{itemize}
    \item O(1) boundary detection: Constant-time wall checking without storing complete maze structure
    \item Compact state space: 50-state Q-table (250 floats: 5 actions $\times$ 50 states) versus full maze memory (e.g., 4,096 cells for 64$\times$64)
    \item Infrequent updates: Q-learning updates every 50 steps reduce per-step overhead
    \item Memory sampling: Sentinel variants reduce visited-cell tracking with 0.06\% performance impact
\end{itemize}

While these design choices reduce memory and update costs, empirical runtime profiling against alternative navigation approaches is not provided.

\subsection{Interpretability}

The agent's decision-making is transparent through:
\begin{itemize}
    \item Coverage percentage: Intuitive progress metric
    \item Manhattan distance: Clear goal proximity measure
    \item Logged switches: Role transitions are recorded and analyzable
    \item Q-value inspection: Learned preferences are queryable
\end{itemize}

This interpretability is valuable for debugging, validation, and deployment in safety-critical applications.

\subsection{Limitations}

The approach has several limitations:

\begin{enumerate}
    \item Single-agent focus: Multi-agent coordination or cooperative exploration is not tested.
    
    \item Discrete action space: Threshold values are discrete (20-60\%); continuous actions may improve performance.
    
    \item Hand-crafted rewards: The reward function is manually designed; learned rewards could be more adaptive.
    
    \item Domain-specific: While the switching mechanism generalizes, the spiral and A* policies are maze-specific.
    
    \item Limited generalization testing: All test mazes share similar structure (grid-based, single target).
    
    \item Computational constraints: Incomplete 128$\times$128 results (86.7\% complete) suggest resource limitations may affect experimental coverage at large scales.
\end{enumerate}

\subsection{Potential Generalization Beyond Mazes}

While demonstrated only on maze navigation, the core principle---learning when to switch between orthogonal policies based on internal progress signals---suggests potential application to domains with similar trade-offs, though adaptation would require domain-specific state representations and action spaces:
\begin{itemize}
    \item Robotics: Switching between area coverage and target approach (analogous progress metrics: coverage, distance)
    \item Search problems: Balancing breadth-first vs. greedy search (analogous signals: explored nodes, goal proximity)
    \item Resource allocation: Transitioning from exploration to exploitation in multi-armed bandits \cite{lattimore2020bandit} (analogous metrics: samples per arm, reward certainty)
    \item Neural architecture search: Switching between architecture exploration and training \cite{zoph2017neural} (analogous signals: architectures tried, validation performance)
\end{itemize}

Empirical validation in these domains remains future work, as the maze-specific components (spiral exploration, A* convergence) would need domain-appropriate replacements.

\section{Related Work}
\label{sec:related}

\subsection{Maze Exploration Algorithms}

Classical maze-solving approaches include wall-following algorithms, depth-first search, breadth-first search \cite{russell2020artificial}, and frontier-based exploration \cite{yamauchi1997frontier}. The sentinel exploration extends these ideas with O(1) boundary detection and systematic coverage guarantees. Prior work has demonstrated the importance of memory in maze navigation \cite{pasukonis2022memorymaze}, which this approach addresses through either full memory (Spiral) or memory sampling (Sentinel).

\subsection{Reinforcement Learning Foundations}

This work builds on foundational Q-learning \cite{watkins1992qlearning} and extends it to the meta-problem of policy selection. The use of epsilon-greedy exploration and tabular Q-learning follows established practices in discrete state spaces \cite{sutton2018reinforcement}. The sparse reward challenge in maze navigation is well-documented \cite{riedmiller2018learning}, which the reward shaping addresses through multi-component objectives.

\subsection{Hierarchical and Options-Based RL}

While this approach involves switching between policies, it differs from hierarchical RL \cite{barto2003options, sutton1999options} in that the focus is on learning \emph{when} to switch rather than constructing a hierarchy of skills. The policies (spiral and A*) are pre-defined rather than learned, focusing the RL problem on timing rather than skill acquisition. This contrasts with options frameworks \cite{sutton1999options} where both options and termination conditions can be learned.

\subsection{Exploration Strategies in RL}

Effective exploration remains a central challenge in reinforcement learning \cite{pathak2017curiosity, burda2019exploration}. This approach addresses exploration through a structured spiral pattern combined with learned switching to exploitation. This differs from curiosity-driven exploration \cite{pathak2017curiosity} which relies on prediction errors, and from count-based methods \cite{bellemare2016unifying} which use visitation frequencies.

\subsection{Policy Composition and Switching}

Recent work on policy composition includes mixture of experts (MoE) architectures \cite{jacobs1991adaptive, shazeer2017outrageously, fedus2022switch} and modular policy learning \cite{andreas2017modular}. While conceptually related, this approach differs in using hand-crafted policies (spiral, A*) rather than learned modules, and focuses on learning discrete switching thresholds rather than soft gating mechanisms. The principle of dynamic policy selection based on state, however, is analogous.

\section{Conclusion}
\label{sec:conclusion}

This work demonstrates that reinforcement learning can effectively learn when to switch between exploration and exploitation strategies in maze navigation. Q-learning with minimal state representation (coverage percentage and distance to goal) enables agents to adapt switching thresholds to problem structure, achieving 23-55\% performance improvements over fixed heuristics across fully tested maze sizes (16$\times$16 through 128$\times$128). Performance gains increase with maze complexity within the tested range, and learned behavior generalizes within each size class to unseen wall configurations while reducing runtime variance by up to 83\%.

The approach requires minimal runtime domain knowledge (maze dimensions and target location), though design choices for state discretization, action space, and reward function embed prior domain insight. The learned policy provides interpretable decision-making through transparent progress signals. Compared to full-memory approaches, O(1) boundary detection and compact Q-tables reduce memory requirements. Ablation studies confirm that both convergence capability and adaptive threshold selection are essential for scalable performance.

The results provide empirical evidence that meta-policy learning---learning when to switch between pre-defined strategies---offers a practical alternative to end-to-end policy learning in domains with identifiable task phases. By decomposing the problem into (1) strategy execution and (2) strategy selection timing, the approach reduces the learning burden while maintaining adaptability. This decomposition may prove valuable in domains where expert strategies exist but their optimal application timing depends on context that cannot be predetermined.

Furthermore, the scaling behavior observed across maze sizes suggests that the value of adaptive switching increases with problem complexity. The 23\% improvement for small mazes grows to 55\% for medium mazes, indicating that as the space of possible problem structures expands, fixed heuristics become increasingly suboptimal relative to learned adaptive policies. This trend, if validated in other domains, would strengthen the case for meta-policy learning in complex, structured environments.

\subsection{Future Directions}

Several promising directions for future work include:

\begin{enumerate}
    \item Continuous action spaces: Using policy gradient methods or actor-critic architectures to learn continuous threshold values instead of discrete 20-60\% increments. This would provide finer-grained adaptation and potentially improve performance, while requiring only algorithmic changes to the existing framework. Proximal Policy Optimization (PPO) \cite{schulman2017proximal} or Soft Actor-Critic (SAC) \cite{haarnoja2018soft} could be particularly well-suited for this continuous control formulation.
    
    \item Cross-episode learning: Implementing Q-table persistence to enable learning across multiple maze runs. Currently, Q-tables are initialized fresh for each episode, but saving and loading learned Q-values could enable transfer learning from small mazes to larger environments, accelerating convergence and reducing per-episode adaptation time. This requires implementing Q-table serialization and developing initialization strategies for transferring knowledge across maze sizes. Meta-learning approaches such as Model-Agnostic Meta-Learning (MAML) \cite{finn2017model} could further enhance cross-episode adaptation speed.
    
    \item Domain generalization: Testing on non-maze environments requiring exploration-exploitation trade-offs (e.g., robotic navigation, search problems). This would validate the broader applicability of adaptive policy switching and requires implementing domain-specific state representations and base policies. Warehouse automation, autonomous vehicle navigation in unknown terrain, and information foraging tasks represent promising testbeds for validating the generality of learned switching policies.
    
    \item Function approximation: Replacing tabular Q-learning with neural networks for richer state representations \cite{mnih2015humanlevel}. This would enable scaling to larger state spaces and more complex features, though it introduces training stability challenges and computational overhead. Deep Q-Networks (DQN) with experience replay and target networks could maintain sample efficiency while handling continuous or high-dimensional state spaces.
    
    \item Learned rewards: Using inverse RL or preference learning to discover optimal reward structures. This would eliminate hand-crafted reward design but requires collecting preference data or demonstrations and implementing meta-learning frameworks. Learning reward functions from human feedback or expert demonstrations could improve adaptability to new task variations without manual reward engineering.
    
    \item Alternative base policies: Investigating other policy pairs beyond spiral exploration and A* pathfinding. For instance, combining frontier-based exploration \cite{yamauchi1997frontier} with rapidly-exploring random trees (RRT) \cite{lavalle1998rapidly} for motion planning, or pairing curiosity-driven exploration \cite{pathak2017curiosity} with model-based planning could reveal whether adaptive switching generalizes across diverse policy combinations.
    
    \item State representation learning: Using autoencoders or variational inference to learn compressed state representations from raw observations rather than hand-crafting coverage and distance features. This would test whether the switching mechanism can operate effectively with learned rather than engineered state spaces, potentially improving generalization to environments where informative features are not known a priori.
    
    \item Hierarchical switching: Extending the two-policy framework to hierarchical structures with multiple levels of strategy selection. For example, a high-level policy could choose between different exploration patterns (spiral, frontier-based, random walk), while a mid-level policy selects between exploration and convergence, and a low-level policy executes primitive actions. This hierarchical decomposition could provide finer-grained control and improved performance in highly complex environments.
\end{enumerate}

The core contribution---using Q-learning to adaptively select when to switch between orthogonal strategies---provides a foundation for more sophisticated approaches to compositional policy learning in structured domains.

\bibliographystyle{IEEEtran}
\bibliography{references}

\end{document}